\documentclass[letterpaper]{article}

\usepackage{natbib,alifeconf}  
\usepackage{booktabs}
\input{preamble}
\usepackage{url,hyperref,cleveref}

\title{Decoding Communications with Partial Information}

\author{
    Dylan Cope$^{1,2}$, \and
    Peter McBurney$^2$ \\
    $^1$University of Oxford, United Kingdom \\
    $^2$King's College London, United Kingdom
} 

\begin{document}

\maketitle

\begin{abstract}
    Machine language acquisition is often presented as a problem of imitation learning: there exists a community of language users from which a learner observes speech acts and attempts to decode the mappings between utterances and situations.
    However, an interesting consideration that is typically unaddressed is partial observability, i.e. the learner is assumed to see all relevant information.
    This paper explores relaxing this assumption, thereby posing a more challenging setting where such information needs to be inferred from knowledge of the environment, the actions taken, and messages sent.
    We see several motivating examples of this problem, demonstrate how they can be solved in a toy setting, and formally explore challenges that arise in more general settings.
    A learning-based algorithm is then presented to perform the decoding of private information to facilitate language acquisition.
\end{abstract}

\section{Introduction}

Consider the problem of an infant watching adults speak and trying to figure out what is being spoken about.
They observe back-and-forth sequences of sounds from each adult, and some globally accessible information granted by the child's embodied perspective.
But the adults may be speaking about any number of topics, or relying on information out-of-sight.
In general, decoding such communications may seem hopeless, but in this paper we explore a constrained form of this problem to illuminate a possible learning mechanism that may aid language acquisition.
Namely, we consider settings in which the `demonstrators' (the adults) are engaged in a grounded, cooperative environment, and the language learner (the infant) has prior knowledge of the structure of the environment.

For each speech act, a \textit{speaker} produces an utterance after having made some observation from the environment.
A \textit{listener} hears this utterance, along with its own observation, and produces some action to forward some goal in the environment.
We will refer to this group of agents as the \textit{target community}, as they are the community using the learner's \textit{target language}. 
In this setting, we will assume that the speaker and listener have access to private information that is not presented to the language learner, and this information is critical for the agents to coordinate in the environment.
The learner thereby observes a dataset $\mathcal{D}$ of communicative interactions between members of the target community, where each sample at time $t$ contains the message sent $m_t$, actions taken by the speaker and listener $a_t^s, a_t^r$ respectively, and any public information $g_t$.
Therefore, the key challenge posed in this paper is to use this information to decode the unobserved (to the learner) observations of the speaker and listener, $o^s_t, o^r_t$.


To address this challenge, we propose to leverage the assumption that agents in the target community are \textit{rational reward maximisers}.
As we will see, combined with prior knowledge of the environment, this can be used to make inferences.
This paper makes the following contributions:
\begin{itemize}[noitemsep]
    \item Posing a novel formal challenge for decoding hidden information from communications.
    \item Formal analysis of this challenge, highlighting key considerations for developing algorithms.
    \item A baseline learning-based algorithm for decoding communications.
\end{itemize}

\section{Goal-Signalling Gridworld Problem}\label{sec:poclap-motivating-examples}

Consider a simple task in which a speaker observes the location on a grid that a listener needs to move to.
This setting is fully cooperative, so the speaker and listener team are rewarded or penalised together.
The episode terminates with a +1 reward when the listener arrives at the goal location, and the team receives a -1 penalty for every time step that the goal is not reached.
The speaker only observes the goal, and the listener observes its location and the message from the speaker.
The listener takes actions to move around the grid, and the speaker sends 4-bit binary messages, i.e.\ $\Sigma = \mathds{Z}^4_2$.
This choice of message space is chosen arbitrarily, and the method we will outline can be applied to any discrete message space.
This is a \textit{referential game}~\citep{lewis_convention_1969} where rather than the listener just recovering the goal location, they also need to navigate to it.


\paragraph{What do your actions say about your words?}
To demonstrate how we may go about decoding the meaning of messages, suppose that rather than only considering a single action at a point in time, the learner collects the data together to look at the sequences of actions that the listener took after receiving a message.
So perhaps the learner observed that the message `0011' was uttered by the speaker, and then the listener took the actions `right', `right', and `up'.
If we now \textit{assume that the agents are rational}, i.e.\ selecting actions and messages to maximise the cooperative reward, we can list the set of possible goals that are consistent with such optimal policies.
Put differently, if the goal were not in one of these locations, an optimal listener would not have taken those actions.

\begin{figure}
    \centering
    \includegraphics[width=\linewidth]{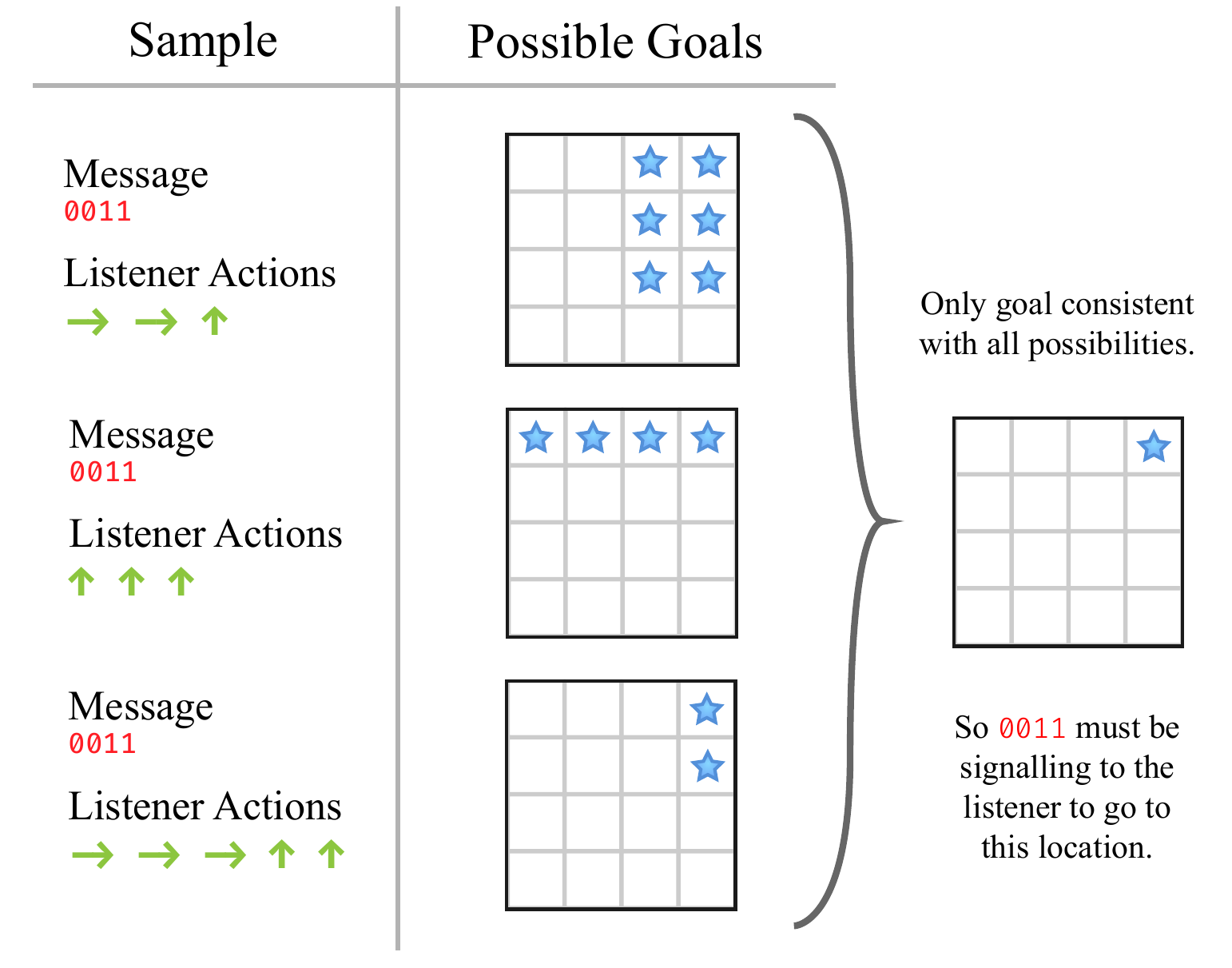}
    \caption{
        Illustration of motivating example.
        The sets of possible goals for each sample are computed by assuming that the agents are rational and selecting actions and messages to maximise the cooperative reward.
    }\label{fig:poclap-example}
\end{figure}

Figure~\ref{fig:poclap-example} demonstrates how this assumption can be used to narrow down the possible goal locations.
For each row of the table in the figure, the left-hand column shows a sample from the dataset that the learner has access to.
The right-hand side shows the set of possible goal locations consistent with the given sequence of actions for an optimal listener.
Looking at the first sample in the figure, we know that the listener would never have taken the `right' action twice unless they had started in one of the two leftmost columns of the gridworld and the goal was in one of the two rightmost columns.
Similarly, we know that the listener would have never taken the `up' action if they had started in the top row, as doing so would cause the agent to remain in place --- a behaviour that gets them no closer to the goal while accumulating time penalty.
Therefore, we can conclude that the agent must have started in one of the tiles in the first two columns and the bottom three rows.
Extrapolating from the possible starting locations with the known actions, we can conclude that the goal must have been in one of the tiles in the two rightmost columns and top three rows.

\section{Background}

\paragraph{Formalising Cooperative Decision-Making.}
A \gls{decpomdp} is a formal model of a cooperative environment defined as a tuple  $\mathcal{M} =(\mathcal{S},\mathcal{A},T,r,\boldsymbol{\Omega},O)$~\cite{oliehoek_concise_2016}, where $\mathcal{S}$ is a set of states, and $\mathcal{A} = \prod_i \mathcal{A}_{i}$ is a product of individual agent action sets. A \textit{joint action} $\mathbf{a} \in \mathcal{A}$ is a tuple of actions from each agent that is used to compute the environment's transition dynamics, defined by a probability distribution over states $T:\mathcal{S}\times\mathcal{A}\times\mathcal{S}\rightarrow[0,1]$. Team performance is defined by a cooperative reward function $r: \mathcal{S} \times \mathcal{A} \times \mathcal{S}$ over state transitions and joint actions. $\boldsymbol{\Omega} = \{\Omega_{i}\}$ is a set of observation sets, and $O:S\rightarrow\prod_i \Omega_i $ is an observation function.

Each agent $i$ follows a policy $\pi_i$ that maps an observation sequence (or a single observation if $i$ is \textit{memoryless}) to a distribution over its actions.
A \textit{trajectory} for an agent $i$ is a sequence of observation-action-reward tuples $\tau_i \in \mathcal{T}_i = (\Omega_i\times\mathcal{A}_i\times\mathds{R})^*$. For a set of policies $\Pi = \{\pi_i\}$, a joint trajectory is $\boldsymbol{\tau} \in \mathcal{T} = (\boldsymbol{\Omega}\times\mathcal{A}\times\mathds{R})^*$.
This paper only considers finite-horizon Dec-POMDPs, so the lengths of trajectories will always be bounded.

\paragraph{Communication Problems.}
Communication is often necessary for agents to coordinate their actions and facilitate cooperation.
We can represent these settings a special kind of \gls{decpomdp}.
The key modification to the standard set-up is that each agent's action set can be expressed as $\mathcal{A}_i = \mathcal{A}_i^e \times \mathcal{A}^c_i$ or $\mathcal{A}_i = \mathcal{A}_i^e \cup \mathcal{A}^c_i$, where $\mathcal{A}_{i}^c$ is a set of \textit{communicative actions}, and $\mathcal{A}_i^e$ is a set of \textit{environment actions}. and agents are not programmed to send messages with any prescribed meaning.
This variant of a \gls{decpomdp} is known as a \gls{comm-decpomdp} \citep{goldman_decentralized_2004,goldman_communication-based_2008,oliehoek_concise_2016}.

In this paper we will only consider situations where the action space is a product ($\mathcal{A}_i^e \times \mathcal{A}^c_i$), meaning that at each time step an agent has the option to both send a message and act in the environment.
The messages have no prior semantics as the transition function of the \gls{decpomdp} only depends on the environment actions ${\mathcal{A}_i^e}$,
The full set of communication symbols is denoted $\Sigma = \bigcup_i \mathcal{A}^c_i$.



\paragraph{Cognitive Science of Language Acquisition.}
Using the assumption of rational speakers has interesting parallels with theories of first language acquisition that draw upon the \emph{intentional stance}.
The intentional stance is concept introduced by the philosopher Daniel Dennett to describe the act of ascribing agentic characteristics to other entities, such as beliefs, desires, intentions, and rationality, in order to predict their behaviour
~\citep{dennett_intentional_1989,dennett_intentional_1971,dennett_intentional_2009}.
Children watching adults around them converse in an unintelligible language, with important context often missing, are effectively placed in their own partially observable language acquisition problems.
\cite{micheal_intentional_2015} argues that the intentional stance plays a role in various kinds of cultural learning, including language acquisition.
In~\cite{gergely_teleological_2003} and~\cite{gergely_rational_2002}, the authors present evidence that children use `rational imitation' and `teleological reasoning' to infer the goals of others.
\cite{bloom_intentionality_1997} discusses the importance of `theory of mind' in how children learn the meanings of novel words, especially nouns.
Bloom further argues that these can be conceived as `conceptual biases about the external world' that aid in language learning~\citep{bloom_capacities_1998}.
A particularly relevant experimental example is the work of~\cite{vouloumanos_twelve-month-old_2012}.
By tracking the gazes of the children, this study showed that 12-month-old infants could recognise when speech communicated the unobserved information.

\section{Strategic Equivalence Classes in Dec-POMDP-Comms}\label{sec:env-strategic-equiv-classes}

Fundamentally, the problem that we are trying to solve is identifying which of the possible optimal policies is the one that the target community of agents are using.
We will investigate the task ahead by first unpacking the structure of the set of possible optimal joint policies $\boldsymbol{\Pi}^*$ for a Dec-POMDP-Comm $\mathcal{M}$.
To make our formalisms concrete, we will use a running example:

\begin{example}\label{ex:goal-comms-strategies-ex}
    Let $\mathcal{M}_{\bigstar}$ be a Dec-POMDP-Comm for a goal-signalling gridworld problem.
\end{example}

There are two important ways to divide up this set of optimal policies $\boldsymbol{\Pi}^*$ for our purposes: (1) policies that act the same way in the environment, and (2), policies that communicate in the same ways.
Firstly, consider that the Dec-POMDP-Comm $\mathcal{M}$ may be solvable in different ways, i.e.\ there could be situations where different optimal policies take different actions.
Formally, there may be optimal joint policies $\boldsymbol{\pi}^1, \boldsymbol{\pi}^2 \in \boldsymbol{\Pi}^*$ and one or more joint observations $\mathbf{o} \in \boldsymbol{\Omega}$ such that $\boldsymbol{\pi}^1(\mathbf{o}) \neq \boldsymbol{\pi}^2(\mathbf{o})$.
In this case, we will say that $\boldsymbol{\pi}^1$ and $\boldsymbol{\pi}^2$ are implementing different \textit{environment-level strategies}.
These could be policies that achieve the same outcomes by different means, or they could be policies that achieve different outcomes of equal value.

\begin{definition}[Environment-level Strategic Equivalence]\label{def:env-strategic-equiv}
    For a Dec-POMDP-Comm $\mathcal{M}$, consider joint policies $\boldsymbol{\pi}_A, \boldsymbol{\pi}_B \in \boldsymbol{\Pi}$.
    These joint policies are environment-level strategically equivalent, denoted $\boldsymbol{\pi}_A \envstratequiv \boldsymbol{\pi}_B$, if $\boldsymbol{\pi}_A(\mathbf{o}) = \boldsymbol{\pi}_B(\mathbf{o})$ for all joint observations $\mathbf{o} \in \boldsymbol{\Omega}$.
\end{definition}

Furthermore, for any joint policy $\boldsymbol{\pi}$, we can define a \emph{strategic equivalence class} $[\boldsymbol{\pi}]$ as the set of all joint policies that implement the same environment-level strategy:

\begin{definition}[Environment-level Strategic Equivalence Class]\label{def:env-strategic-equiv-class}
    Given a Dec-POMDP-Comm $\mathcal{M}$ and joint policy $\boldsymbol{\pi} \in \boldsymbol{\Pi}$, the environment-level strategic equivalence class of $\boldsymbol{\pi}$ is the set of all joint policies that implement the same environment-level strategy:
    \begin{align}
        {[\boldsymbol{\pi}]}^e = \left\{\boldsymbol{\pi}' \in \boldsymbol{\Pi}~|~\boldsymbol{\pi}' \envstratequiv \boldsymbol{\pi}\right\}
    \end{align}
\end{definition}

\begin{figure}
    \centering
    \includegraphics[width=.95\linewidth]{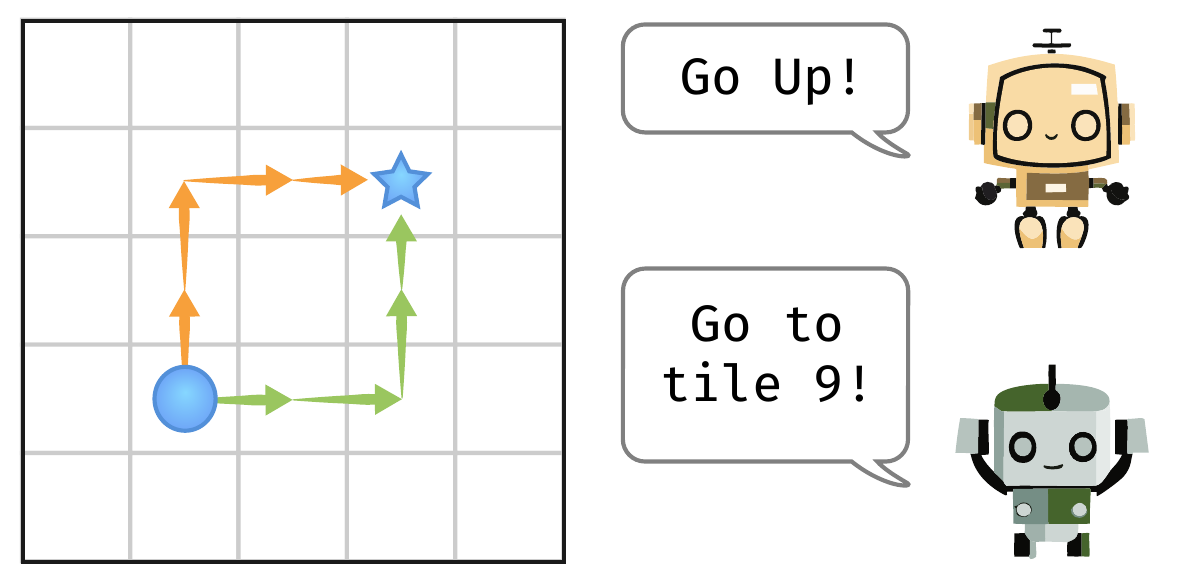}
    \caption[Illustration of different environment-level and communication-level strategies for a gridworld goal communications problem.]{
        Illustration of different environment-level and communication-level strategies for a goal-signalling problem.
        In the gridworld, we see the listener (blue circle) and the goal (blue star), and two equally-optimal trajectories to reach the goal depicted with green and orange arrows.
        On the right, outside the grid, we see two speakers employing different equally-optimal communication strategies.
    }\label{fig:dec-pomdp-strategy-class-illustration}
\end{figure}

\begin{continueexample}{ex:goal-comms-strategies-ex}
    For $\mathcal{M}_{\bigstar}$ we see the trajectories for two optimal environment-level strategic equivalence classes depicted with green and orange arrows in Figure~\ref{fig:dec-pomdp-strategy-class-illustration}.
    Note that from the same starting locations, these different strategies take different actions.
    If we consider how each of these environment-level strategies can be implemented with different communication strategies, we see that the speaker who is saying `Go Up!' is only compatible with the orange strategy.
    On the other hand, the speaker who is saying `Go to tile 9!' is compatible with both strategies.
\end{continueexample}

\begin{definition}[Optimal Strategy Set]\label{def:optimal-strategy-set}
    Given a Dec-POMDP-Comm $\mathcal{M}$, the set of all optimal environment-level strategic equivalence classes is:
    \begin{align}
        \text{Optimal-Strategies}(\mathcal{M}) = \left\{{[\boldsymbol{\pi}]}^e~|~\boldsymbol{\pi} \in \boldsymbol{\Pi}^*\right\}
    \end{align}

\end{definition}

\begin{theorem}
We can express the set of all optimal policies as a union of the optimal environment-level strategic equivalence classes:
\begin{align}
    \boldsymbol{\Pi}^* = \bigcup_{\boldsymbol{\Pi} \in S} \boldsymbol{\Pi},\quad\text{where}~S = \text{Optimal-Strategies}(\mathcal{M})
\end{align}
\end{theorem}

In a typical \gls{decpomdp} without communication, if two policies $\boldsymbol{\pi}^1, \boldsymbol{\pi}^2$ satisfy $\boldsymbol{\pi}^1(\mathbf{o}) = \boldsymbol{\pi}^2(\mathbf{o})$ for all joint observations $\mathbf{o}$, then they are just the same policy.
This brings us to the second important way to divide up $\boldsymbol{\Pi}^*$.
Because each agent's policies $\boldsymbol{\pi} = (\pi_A, \pi_B)$ in a Dec-POMDP-Comm can be factored into environment-level and communication-level policies $\pi_1 = (\pi^e_A, \pi^c_B)$, joint policies in the same environment-level strategic equivalence class can have different communication policies.
More precisely, we can define an equivalence relation on the set of joint policies $\boldsymbol{\Pi}$ for teams using the same communication strategy:


\begin{definition}[Communication-level Strategic Equivalence]\label{def:comm-strategic-equiv}
    For a Dec-POMDP-Comm $\mathcal{M}$, consider two environment-level strategically equivalent joint policies $\boldsymbol{\pi}_A \envstratequiv \boldsymbol{\pi}_B \in \boldsymbol{\Pi}$.
    Each joint policy is composed of $N$ agents $\boldsymbol{\pi}_x = (\pi_{x,1}, \ldots, \pi_{x,N})$.
    Each agent's policy can be factored into environment and communication policies, i.e.\ $\pi_{x, i} = (\pi^e_{x, i}, \pi^c_{x, i})$, where the communication policy is a mapping from the agent's observations to message space $\pi^c_{x, i}: \Omega_i \to \Sigma_i$.
    Consider agent $i$, the communication policies $\pi^c_{A, i}$ and $\pi^c_{B, i}$ are strategically equivalent if there exists a bijection $\phi_i: \Sigma_i \to \Sigma_i$ between their images,\ i.e.:
    \begin{align}
        \pi^c_{A, i}(o_i) = \phi_i(\pi^c_{B, i}(o_i))\quad\forall o_i \in \Omega_i
    \end{align}
    Therefore, the joint policies are communication-level strategically equivalent, denoted $\boldsymbol{\pi}_A \commstratequiv \boldsymbol{\pi}_B$, if $\pi^c_{A, i}$ and $\pi^c_{B, i}$ are strategically equivalent for all agents $i$.

\end{definition}

By the same means as Definition~\ref{def:env-strategic-equiv-class}, we can define the set of all optimal communication-level strategic classes for a Dec-POMDP-Comm $\mathcal{M}$:

\begin{definition}[Communication-level Strategic Equivalence Class]\label{def:comm-strategic-equiv-class}
    Given a Dec-POMDP-Comm $\mathcal{M}$ and joint policy $\boldsymbol{\pi} \in \boldsymbol{\Pi}$, the communication-level strategic equivalence class of $\boldsymbol{\pi}$ is the set of all joint policies that implement the same communication-level strategy:
    \begin{align}
        {[\boldsymbol{\pi}]}^c = \left\{\boldsymbol{\pi}' \in \boldsymbol{\Pi}~|~\boldsymbol{\pi}' \commstratequiv \boldsymbol{\pi}\right\}
    \end{align}
\end{definition}

\begin{continueexample}{ex:goal-comms-strategies-ex}
    To understand the role of the bijection in the definition of communication-level strategic equivalence.
    Let us suppose that in this game, there are 30 possible utterances that the speaker could make, which we can write as $|\Sigma| = 30$.
    The top speaker, who is saying `Go Up!', in Figure~\ref{fig:dec-pomdp-strategy-class-illustration} is only using 4 of these; one for each of the cardinal directions that the listener could move in.
    We can denote the set of these messages as $\Sigma^{\text{top}} \subset \Sigma$ and thus $|\Sigma^{\text{top}}| = 4$.
    On the other hand, the bottom speaker, who is saying `Go to tile 9!', has a unique message for each of the 25 possible goal locations, i.e. $|\Sigma^{\text{bottom}}| = 25$.
    Therefore, for these two communication policies to be strategically equivalent, there would need to exist a bijection $\phi: \Sigma^{\text{top}} \to \Sigma^{\text{bottom}}$.
    However, as $|\Sigma^{\text{top}}| \neq |\Sigma^{\text{bottom}}|$, such a bijection cannot exist.





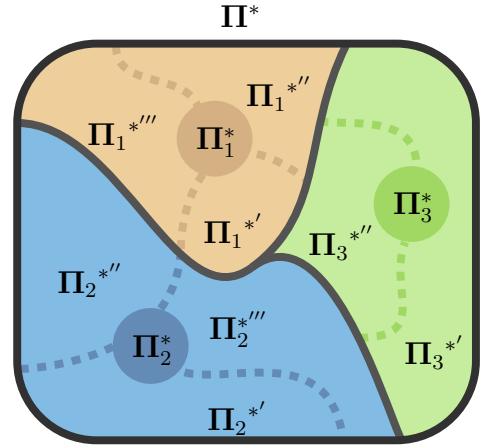
\begin{figure}
    \centering
    \tikzset{every picture/.style={line width=0.75pt}} 

\begin{tikzpicture}[x=0.75pt,y=0.75pt,yscale=-1,xscale=1]

\draw  [draw opacity=0][fill={rgb, 255:red, 130; green, 187; blue, 228 }  ,fill opacity=1 ][line width=3]  (45.5,86.56) .. controls (45.5,64.45) and (63.43,46.52) .. (85.54,46.52) -- (236.8,46.52) .. controls (258.92,46.52) and (276.84,64.45) .. (276.84,86.56) -- (276.84,206.7) .. controls (276.84,228.81) and (258.92,246.74) .. (236.8,246.74) -- (85.54,246.74) .. controls (63.43,246.74) and (45.5,228.81) .. (45.5,206.7) -- cycle ;
\draw  [draw opacity=0][fill={rgb, 255:red, 198; green, 236; blue, 158 }  ,fill opacity=1 ] (181.42,154.18) -- (170.41,155.82) -- (180.01,143.64) -- (190.32,123.95) -- (201.57,73.57) -- (211.93,46.24) -- (249.85,48.03) -- (268.59,61.15) -- (277.03,81.3) -- (277.5,199.18) -- (275.86,216.75) -- (268.36,232.46) -- (254.53,242.77) -- (239.54,247.22) -- (222.43,205.04) -- (207.9,177.15) -- (193.61,160.04) -- cycle ;
\draw [color={rgb, 255:red, 161; green, 216; blue, 99 }  ,draw opacity=1 ][line width=3]  [dash pattern={on 3.38pt off 3.27pt}]  (246.47,114.01) .. controls (250.97,81.39) and (215.79,88.08) .. (198.92,85.83) ;
\draw [color={rgb, 255:red, 99; green, 139; blue, 182 }  ,draw opacity=1 ][line width=3]  [dash pattern={on 3.38pt off 3.27pt}]  (122.36,208.01) .. controls (137.99,226.13) and (202.04,189.26) .. (209.85,246.13) ;
\draw  [draw opacity=0][fill={rgb, 255:red, 238; green, 206; blue, 155 }  ,fill opacity=1 ] (85.54,46.52) .. controls (90.98,45.96) and (123.16,44.36) .. (153.86,45.77) .. controls (184.56,47.17) and (219.43,44.6) .. (211.23,47.64) .. controls (203.03,50.69) and (194.54,135.91) .. (171.11,152.31) .. controls (147.67,168.71) and (143.46,174.1) .. (106.9,129.34) .. controls (70.34,84.58) and (48.55,87.16) .. (47.61,86.69) .. controls (46.67,86.22) and (43.39,86.22) .. (48.78,69.59) .. controls (54.17,52.95) and (80.11,47.08) .. (85.54,46.52) -- cycle ;
\draw [color={rgb, 255:red, 212; green, 176; blue, 131 }  ,draw opacity=1 ][line width=3]  [dash pattern={on 3.38pt off 3.27pt}]  (127.99,150.2) .. controls (127.99,130.2) and (140.67,75.71) .. (192.04,115.83) ;
\draw [color={rgb, 255:red, 83; green, 83; blue, 83 }  ,draw opacity=1 ][line width=3]    (163.98,159.38) .. controls (201.48,131.26) and (189.11,92.64) .. (211.23,47.64) ;
\draw  [draw opacity=0][fill={rgb, 255:red, 99; green, 139; blue, 182 }  ,fill opacity=1 ] (93.67,199.12) .. controls (93.67,188.55) and (102.24,179.99) .. (112.8,179.99) .. controls (123.36,179.99) and (131.93,188.55) .. (131.93,199.12) .. controls (131.93,209.68) and (123.36,218.25) .. (112.8,218.25) .. controls (102.24,218.25) and (93.67,209.68) .. (93.67,199.12) -- cycle ;
\draw [color={rgb, 255:red, 99; green, 139; blue, 182 }  ,draw opacity=1 ][line width=3]  [dash pattern={on 3.38pt off 3.27pt}]  (46.75,210.82) .. controls (86.43,208.32) and (124.24,191.76) .. (127.68,153.95) ;
\draw  [draw opacity=0][fill={rgb, 255:red, 212; green, 176; blue, 131 }  ,fill opacity=1 ] (125.92,94.51) .. controls (125.92,83.94) and (134.48,75.38) .. (145.04,75.38) .. controls (155.61,75.38) and (164.17,83.94) .. (164.17,94.51) .. controls (164.17,105.07) and (155.61,113.63) .. (145.04,113.63) .. controls (134.48,113.63) and (125.92,105.07) .. (125.92,94.51) -- cycle ;
\draw  [draw opacity=0][fill={rgb, 255:red, 161; green, 216; blue, 99 }  ,fill opacity=1 ] (225.28,127.5) .. controls (225.28,116.94) and (233.84,108.38) .. (244.41,108.38) .. controls (254.97,108.38) and (263.53,116.94) .. (263.53,127.5) .. controls (263.53,138.07) and (254.97,146.63) .. (244.41,146.63) .. controls (233.84,146.63) and (225.28,138.07) .. (225.28,127.5) -- cycle ;
\draw [color={rgb, 255:red, 161; green, 216; blue, 99 }  ,draw opacity=1 ][line width=3]  [dash pattern={on 3.38pt off 3.27pt}]  (217.98,195) .. controls (254.72,194.62) and (232.04,170.2) .. (241.1,146.76) ;
\draw [color={rgb, 255:red, 212; green, 176; blue, 131 }  ,draw opacity=1 ][line width=3]  [dash pattern={on 3.38pt off 3.27pt}]  (137.36,89.26) .. controls (141.86,56.64) and (95.37,73.14) .. (95.37,48.02) ;
\draw [color={rgb, 255:red, 83; green, 83; blue, 83 }  ,draw opacity=1 ][line width=3]    (45.5,86.56) .. controls (97.99,87.01) and (126.49,187.5) .. (163.98,159.38) ;
\draw [color={rgb, 255:red, 83; green, 83; blue, 83 }  ,draw opacity=1 ][line width=3]    (163.98,159.38) .. controls (201.48,131.26) and (229.6,224.99) .. (238.97,245.99) ;
\draw  [color={rgb, 255:red, 49; green, 49; blue, 49 }  ,draw opacity=1 ][line width=3]  (45.5,86.56) .. controls (45.5,64.45) and (63.43,46.52) .. (85.54,46.52) -- (236.8,46.52) .. controls (258.92,46.52) and (276.84,64.45) .. (276.84,86.56) -- (276.84,206.7) .. controls (276.84,228.81) and (258.92,246.74) .. (236.8,246.74) -- (85.54,246.74) .. controls (63.43,246.74) and (45.5,228.81) .. (45.5,206.7) -- cycle ;

\draw (146.6,25.6) node [anchor=north west][inner sep=0.75pt]  [font=\Large]  {$\boldsymbol{\Pi }^{*}$};
\draw (134.21,87.47) node [anchor=north west][inner sep=0.75pt] [font=\Large]   {$\boldsymbol{\Pi }_{1}^{*}$};
\draw (101.96,192.08) node [anchor=north west][inner sep=0.75pt]  [font=\Large]  {$\boldsymbol{\Pi }_{2}^{*}$};
\draw (233.57,120.09) node [anchor=north west][inner sep=0.75pt]  [font=\Large]  {$\boldsymbol{\Pi }_{3}^{*}$};
\draw (137.89,130.68) node [anchor=north west][inner sep=0.75pt]  [font=\Large]  {${\boldsymbol{\Pi }_{1}}^{*'} $};
\draw (159.48,61.69) node [anchor=north west][inner sep=0.75pt]  [font=\Large]  {${\boldsymbol{\Pi }_{1}}^{*''} $};
\draw (79.49,81.18) node [anchor=north west][inner sep=0.75pt]  [font=\Large]  {${\boldsymbol{\Pi }_{1}}^{*'''} $};
\draw (139.89,225.29) node [anchor=north west][inner sep=0.75pt]  [font=\Large]  {${\boldsymbol{\Pi }_{2}}^{*'} $};
\draw (64.25,155.8) node [anchor=north west][inner sep=0.75pt]  [font=\Large]  {${\boldsymbol{\Pi }_{2}}^{*''} $};
\draw (239.32,191.98) node [anchor=north west][inner sep=0.75pt]  [font=\Large]  {${\boldsymbol{\Pi }_{3}}^{*'} $};
\draw (190.93,136.68) node [anchor=north west][inner sep=0.75pt]  [font=\Large]  {${\boldsymbol{\Pi }_{3}}^{*''}$};
\draw (140.86,180.16) node [anchor=north west][inner sep=0.75pt]  [font=\Large]  {$\boldsymbol{\Pi }^{*'''}_{2} $};

\end{tikzpicture}
    \caption[Abstract representation of a decomposition of the set of possible optimal policies based on communication and environmental equivalence classes.]{
        Abstract representation of a decomposition of the set of possible optimal policies $\boldsymbol{\Pi}^*$ for a Dec-POMDP-Comm.
        Here we have illustrated three environment-level strategic equivalence classes, denoted by coloured regions of the space.
        Each of these classes is further comprised of communication-level strategic equivalence classes, with these regions demarcated by the dashed lines.
    }\label{fig:dec-pomdp-optimal-policies}
\end{figure}

\paragraph{Implications for goal inference.}
As we saw in the section introducing the goal-signalling gridworld, the assumption that the agents are rational is crucial for decoding the communication protocol when we only observe their messages and actions.
For each sample in the data shown in Figure~\ref{fig:poclap-example}, the method was composed of the following steps:
\begin{enumerate}[noitemsep]
    \item Iterate through each possible observation that the listener could have made (i.e.\ each location they could be in).
    \item If any optimal policy would take the observed actions from this state, then the location of the goal can be inferred. Add this to a set of possible goal locations consistent with the sample.
    \item Take the intersection of the sets of possible goal locations for each sample to find the set of goal locations that are consistent with all the samples.
\end{enumerate}
The set of optimal policies considered in step 2 does not need to be complete, but it needs to include a policy from the correct environment-level strategic equivalence class (Definition~\ref{def:env-strategic-equiv-class}).
This presents an opportunity: if we can narrow down the set of optimal policies that we consider, we can reduce the amount of computation required to perform step 2.
Indeed, determining if `any optimal policy' takes the observed actions may be computationally infeasible if the set of optimal policies is large.
But on the other hand, if we restrict the set of possible optimal policies that we consider, we risk excluding the true policy, and thereby we cannot perform the inference.

\end{continueexample}

\section{Learning to Decode Messages}
\begin{figure*}
    \centering
    \includegraphics[width=.7\linewidth]{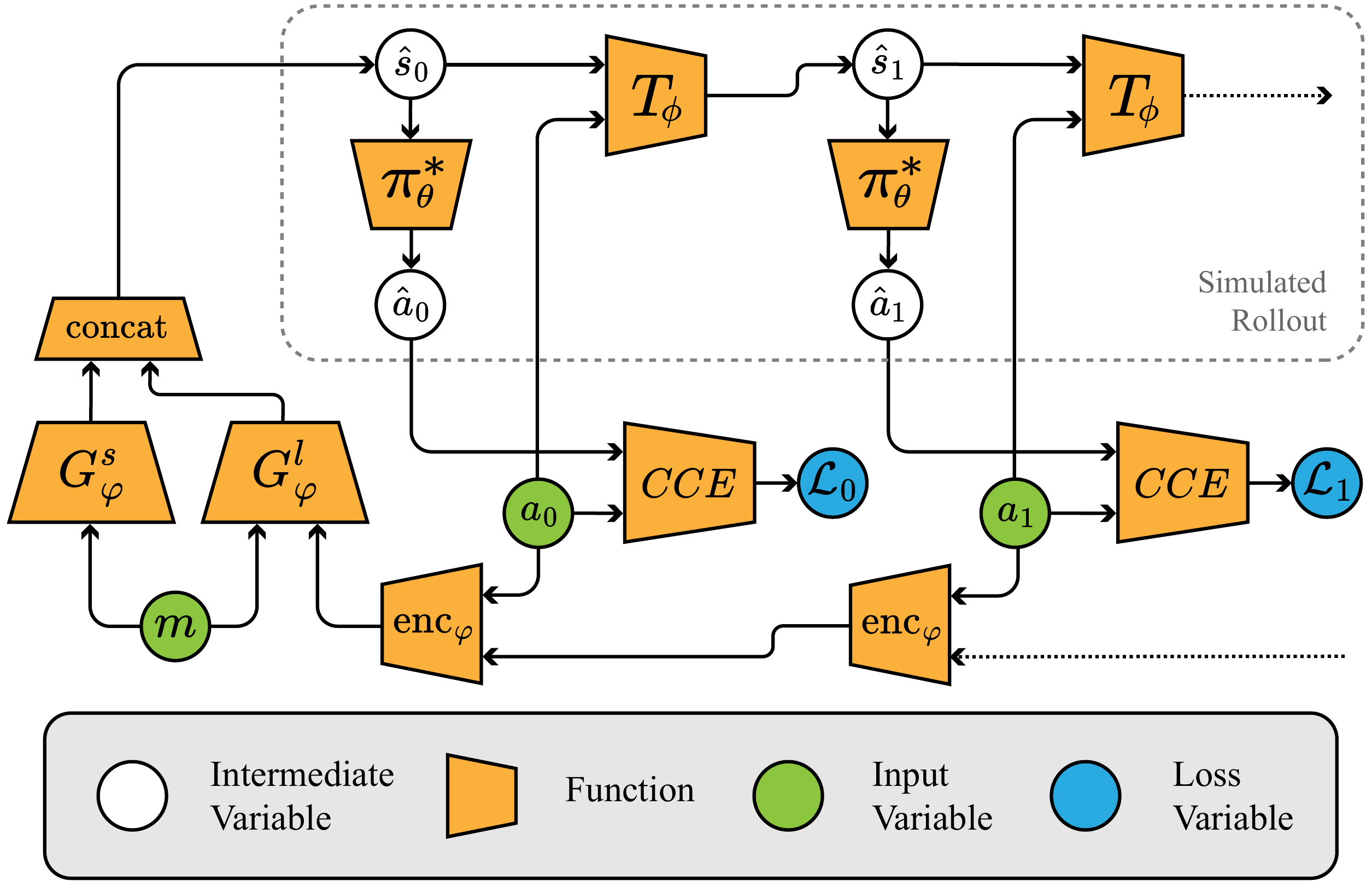}
    \caption[Diagram of the computational graph used for the training the state decoder.]{
        Diagram of the computational graph used for the training the state decoder, which is composed of the models $G^{s}_\varphi$, $G^{l}_\varphi$, and $\text{enc}_\varphi$.
        Arrows indicate the flow of information and the parameters $\varphi$ optimised by backpropagating from the loss variables $\mathcal{L}_t$ (blue) to the message $m$ and action $a_t$ input variables (green).
    }\label{fig:poclap-rollout-train-diagram}
\end{figure*}

\subsection{Method}

This section presents a learning algorithm for solving the hidden-information decoding problem, which we will call the \textit{state decoding} problem.
There are some encouraging initial results in the goal-signalling gridworld problem, but there are clearly some limitations and future work.
The basic architecture of the approach is shown in Figure~\ref{fig:poclap-rollout-train-diagram}.
The algorithm involves separately training three components:
\begin{itemize}
    \item \textbf{Joint Policy:} A policy that maps joint observations to joint actions.
    In the case of the gridworld discussed throughout this paper, the speaker does not have any environment level actions.
    Therefore, we learn a policy that maps the speaker and listener observations to the listener actions.
    We denote this $\pi^*_\theta$ as after training, it will be the optimal policy.
    \item\textbf{Transition Model:} A model $T_\phi(s_t, a_t) = s_{t+1}$ that predicts the next state $s_{t+1}$ given the current state and the joint actions of the agents.
    \item \textbf{State Decoder:} A model that recovers the state of the game from the messages and actions of the agents.
\end{itemize}

The first step is to train the joint policy $\pi^*_\theta$.
As we will later need to backpropagate through the joint policy, we need to use a differentiable policy.
\gls{ppo}~\citep{schulman_proximal_2017} was used to train $\pi^*_\theta$ --- for this problem, the joint policy is effectively a single agent.

Next, to train the transition model $T_\phi$, we need to generate a dataset of transitions and define a loss function over the observation space.
For more complex, high-dimensional observation spaces like images, we would need a more sophisticated loss function, but for this setting the joint observation space is composed of four discrete variables represented as one-hot feature vectors.
Therefore, $T_\phi$ is trained using the sum of cross-entropy losses between the predicted next feature vectors and the true next feature vectors.

Note that the joint observation space is also the state space of this \gls{comm-decpomdp}, so these terms can be used interchangeably.
This is why we call the third model the `State Decoder' rather than `Joint-Observation Decoder', and in Figure~\ref{fig:poclap-rollout-train-diagram} we show the predicted states $\hat{s}_t$ feeding into both the transition model and the joint policy.
Finally, the most complex training process is training the state decoder.
The state decoder is composed of three parts:
\begin{itemize}
    \item \textbf{Actions Encoder:} A model that embeds the observed sequence of actions into a latent space, $\text{enc}_\varphi$ in Figure~\ref{fig:poclap-rollout-train-diagram}.
    \item \textbf{Initial State Generator:} This component comprises two models, one that generates the initial observation of the speaker $G^s_\varphi$, and another that generates the initial observation of the listener $G^l_\varphi$. These outputs of these models are concatenated to produce the first state $\hat{s}_0$.
    \item \textbf{Simulated Rollout:} To produce the sequence of predicted states after the initialisation, $\hat{s}_1, \ldots, \hat{s}_T$, we simulate a game using the joint policy $\pi^*_\theta$ and the transition model $T_\phi$.
    To produce the next state, we use the previous predicted state and the ground-truth action $a_t$.
    After each step, $T_\phi$ outputs logits for the next state, which are used to sample from a Gumbel-Softmax distribution.
\end{itemize}

To train the system we optimise the parameters $\varphi$, leaving the parameters of the joint policy $\theta$ and transition model $\phi$ fixed.
The data for training the state-action decoder is a set of demonstrations of the form $(m, a_0, \ldots, a_{L})$, where $m$ is the message, $a_t$ is the action taken at time $t$, and $L$ is the length of the demonstration.
The formal process for computing the predicted states, actions, and \emph{action-reconstruction loss} is as follows:
\begin{enumerate}
    \item First, process the sequence of actions through the actions-encoder \gls{rnn}, $\text{enc}_\varphi$ to get a latent representation $e_a = \text{enc}_\varphi(a_0, \ldots, a_{L})$.
    \item Next, generate the initial observations for the speaker and listener, $\hat{o}^s_0, \hat{o}^l_0$, using the initial state generator models $G^s_\varphi$ and $G^l_\varphi$: $\hat{o}^s_0 = G^s_\varphi(m)$ and $\hat{o}^l_0 = G^l_\varphi(m, e_a)$.
    \item Use these outputs as logits to sample the set of categorical variables $V_{\text{feats}}$ that encode the initial state:
    \begin{align}
        V_{\text{feats}} \sim \text{Gumbel-Softmax}\left(\hat{o}^l_0, \hat{o}^s_0, \tau\right)
    \end{align}
    The hyperparameter $\tau$ is the temperature of the Gumbel-Softmax distribution~\citep{jang_categorical_2017,maddison_concrete_2017}.
    In the case of the gridworld, the state is composed of four discrete variables, so $V_{\text{feats}}$ is a set of four one-hot feature vectors, $\text{goal}_x, \text{goal}_y, \text{listener}_x, \text{listener}_y$.
    The predicted state is then the concatenation of these vectors: $\hat{s}_0 = \text{concat}(V_{\text{feats}})$.
    \item Simulate a game using the joint policy $\pi^*_\theta$, the demonstration actions $a^i_0,\ldots,a^i_{T}$, and the transition model $T_\phi$ to get the sequence of predicted states $\hat{s}_1, \ldots, \hat{s}_{L}$. For each step $t$, the predicted state and action logits are computed as:
    \begin{align}
        \hat{s}_{t+1} &\sim \text{Gumbel-Softmax}\left(T_\phi(\hat{s}_t, a_t), \tau\right) \\
        a^\text{logits}_{t+1} &= \pi^*_\theta(\hat{s}_{t+1})
    \end{align}
    The same temperature $\tau$ is used as in the initial state generation.
    \item The action-reconstruction loss is computed as the sum of the categorical cross-entropy losses between the predicted actions and the true actions:
    \begin{align}
        \mathcal{L}^{action-reconstr} = \sum_{t<L} \text{CCE}(a^\text{logits}_t, a_t)
    \end{align}
\end{enumerate}

Thereby, the state decoder is trained to predict the state of the game from the messages and actions of the agents.
A critical component of this architecture is separating generation of the speaker and listener observations.
Using one model to generate both observations does not work as the speaker's observation is generated from the message alone, while the listener's observation is generated from the message and the action sequence.

\subsection{Empirical Evaluations}

\begin{figure*}
    \centering
    \includegraphics[width=\linewidth]{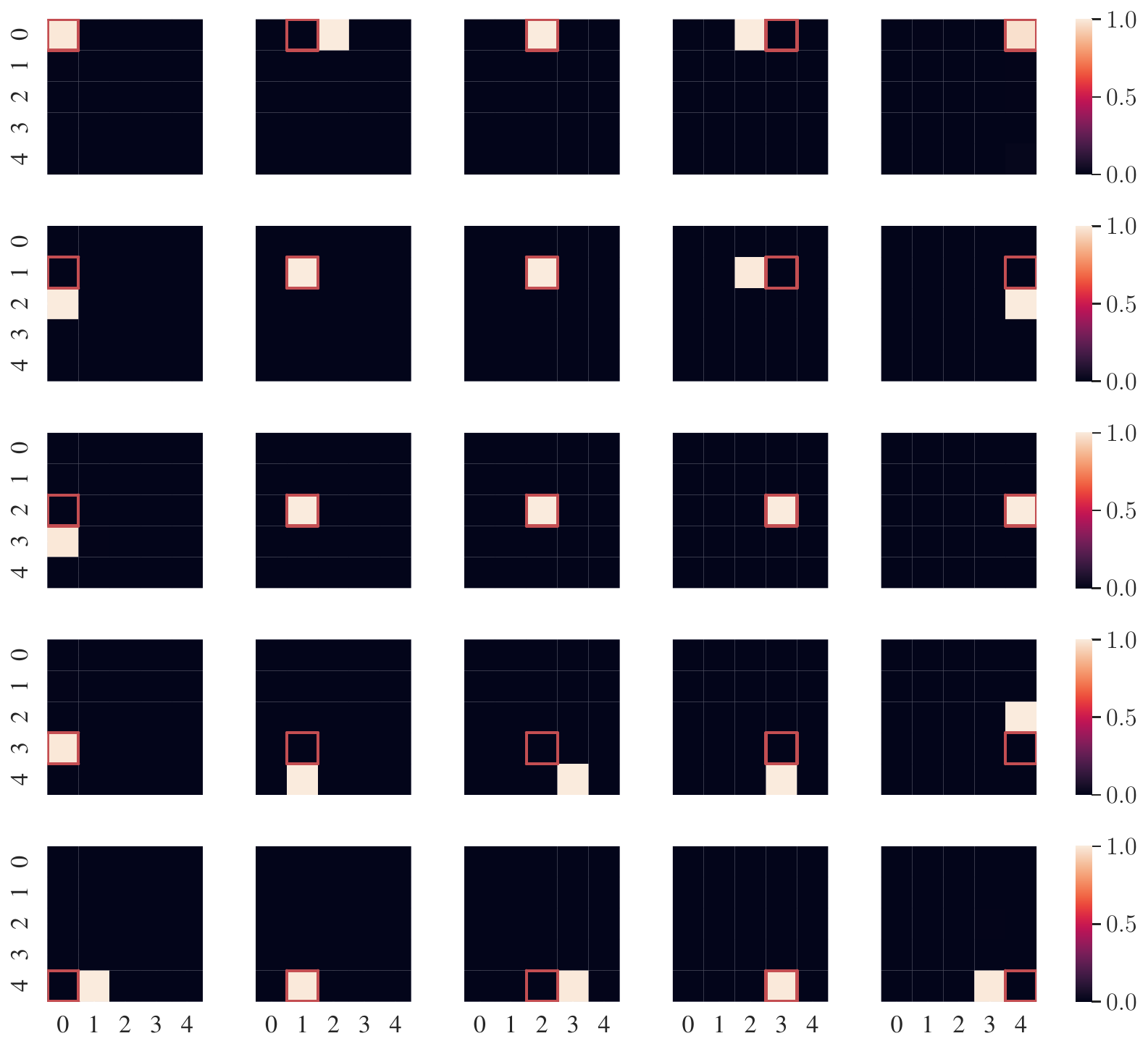}
    \caption[Analysis of the goal position predictions for the state decoder trained to recover the state from any message.]{
        Analysis of the goal position predictions for the state decoder trained to recover the state from any message.
        The heatmap at each $(i, j)$ position in this $5\times 5$ grid of heatmaps corresponds to evaluations where the true goal is at $(i, j)$.
        The heatmaps themselves are also $5\times 5$ grids, where the value at each $(x, y)$ position corresponds to the proportion of evaluations where the predicted goal is at $(x, y)$, normalised between zero and one.
        The $(i, j)$ coordinate is highlighted with a red box to indicate the true goal location.
        Therefore, perfect performance is illustrated by a white square at the true goal location and black squares elsewhere.
        We see this is the case for 12 of the 25 goal locations.
        For 13 cases, the model does not predict the correct locations.
        But in 12 cases it always chooses a location within one step of the true goal (Manhattan distance of one), and in the remaining case it chooses a location within two steps.
    }\label{fig:poclap-goal-predictions-tables}
\end{figure*}

\paragraph{Training the Joint-Policy.}
To construct an optimal joint policy for state decoding, $\pi^*_\theta$ was trained using the PureJAXRL~\citep{lu_discovered_2022} implementation of \gls{ppo} on the goal-signalling gridworld problem.
Within 200 iterations, the joint policy converged to a high reward.

\paragraph{Training the Transition Model.}
To train the transition model $T_\phi$, a dataset of state-action, next-state pairs was generated by simulating the game using the trained $\pi^*_\theta$.
These form a supervised learning problem, and the transition model was trained using the Adam optimiser with a learning rate of $1\times 10^{-3}$.
The loss converged to zero over 300 training steps for the transition model.
To evaluate this model, we can measure the accuracy of its predicted state sequences when applied recursively on its own predictions.
We find the model achieves 100\% accuracy, which is important because as shown in Figure~\ref{fig:poclap-rollout-train-diagram}, the state decoder relying on such `simulated rollouts' to train.
Therefore, an unreliable transition model would lead to unreliable training of the state decoder.


\paragraph{Generating Demonstrations.}
For these tests, we use a greedy policy derived from the trained $\pi^*_\theta$ to generate demonstrators for the learner.
As the policy learned via \gls{ppo} is stochastic, we can sample from it with different temperatures to generate different sets of demonstrations.
Therefore, we can think of the stochastic policy that we have access to during state decoding as a defining a probability distribution over the different environment-level equivalence classes.
By using the same policy to generate the demonstrations, we are assuming not only that the true policy (the one we are trying to decode) is in the set of policies that we sample from, but also that it is the most likely policy in this set.
Finally, messages for the demonstrations were generated synthetically by assigning each goal location a unique message based on an arbitrary mapping.
This mapping is fixed throughout each experiment, so there is always a consistent relationship between messages and goal locations to decode.


\paragraph{Training the State Decoder.}
The state decoder is trained using the Adam optimiser with a learning rate of $1\times 10^{-3}$, optimising a loss function that is the sum of the cross-entropy losses between the predicted actions and the true actions.
At each training step, a batch of 512 demonstration episodes is generated.
During training, 512 environments are initialised and stepped 8 times.
Sequences where the listener is initialised to the same location as the goal are discarded, as the episode is immediately terminated.
Likewise, sequences where the episode does not end are also discarded.
To evaluate the performance of the state decoder, we can look at the accuracy of the goal positions: the frequency that the predicted position equals the actual goal.



An exponential temperature schedule was used for the Gumbel-Softmax samples.
The schedule starts at a temperature of 10.0 and decays to 0.5 over 15000 training steps, with a full training run of 20000 training steps.
The temperature is updated 500 steps.
The loss dropped rapidly at first, plateaus for a while, then drops again to a minimum after around 10000 training steps.
By this point, the goal prediction accuracy converged to around 50\%, where it remains for the rest of training despite the loss increasing slightly, and occasionally spiking.
This effect on the loss is likely due to the temperature schedule, which is still decaying at this point.


We can visualise the predictions of the state decoder by constructing a table of heatmaps, shown in Figure~\ref{fig:poclap-goal-predictions-tables}.
In each set of evaluations, the model is given sequences of actions and messages, and it predicts the goal locations.
Each heatmap corresponds to the predictions made by the model when the true goal is at the location corresponding to the position of the heatmap in the table.
As the heatmaps are normalised, the value at each position corresponds to the proportion of evaluations where the predicted goal is at that location.
We see that in all cases, the state decoder has converged on a single answer corresponding to the true goal location.
In 12 of the 25 cases, the model predicts the correct location.
In every other case, the model predicts a location within one step of the true goal, except for one case where it predicts a location within two steps.
This indicates that even in the cases where the model is wrong, it has uncovered a form of semantic similarity between messages. 

\renewcommand{\thefootnote}{\fnsymbol{footnote}}
\footnotetext{The code for the experiments in this paper can be found at: \url{github.com/DylanCope/decoding-communications}}







\section{Related Work}

The idea of machine language acquisition has been approached from a variety of perspectives.
Most prominently in recent years, \glspl{llm} have emerged as a dominant approach.
Effectively, this is straight-forward imitation learning in the form of behavioural cloning~\citep{widrow_pattern_1964,sammut_behavioral_2010,hussein_imitation_2017}.
Thus, \glspl{llm} suffer from sensitivity to the demonstration data~\cite{kumar_should_2022}.
Additionally, they are unable to ground words in real-world actions, and no consideration is given to private information of speakers.

In emergent communication research, systems of agents develop communication systems, whether through evolutionary models of signalling and language emergence~\cite{,ackley_altruism_1994,bullock_evolutionary_1997,parisi_artificial_1997,mirolli_evolving_2010}, large-scale robotic experiments such as Steels’ Talking Heads~\citep{steels_talking_1999}, or more recent neural approaches~\citep{lazaridou_emergent_2020,havrylov_emergence_2017,wagner_progress_2003,chaabouni_emergent_2022,foerster_learning_2016}, with a variety of approaches and settings considered.
Compared to language modelling, this line research confronts the issue of grounded language, however, agents develop their own communication systems, rather than learning an existing language.
A step in that direction is the idea of iterated learning and simulations of cultural transmission~\citep{smith_iterated_2003,kirby_emergence_2002,kirby_cumulative_2008}.
These works primarily focus on how an evolving population of language users, with new agents continuously being added, applies pressure on the emergence of certain kinds of linguistic structures (e.g. compositionality).

At the intersection of imitation learning and emergent communication, \glspl{clap} have been proposed as a framework for learning the grounded communication system of a target community of language users~\citep{cope_joining_2022,cope_learning_2024}.
Similarly to this paper, \cite{cope_learning_2024} looked at learning from a dataset of speech-act demonstrations with action and observations, thereby requiring a more privileged position for the learning agent by assuming full visibility.
Additionally, the proposed algorithms in that work are aimed at learning policies, while the state decoding algorithm proposed here is just concerned with reconstructing the hidden information.
This presents an opportunity for future work combining the algorithms into a larger system for solving \glspl{clap} with partial observability for the learner.

Another related research area is the recent interest in the emergence of \textit{covert signalling}, especially in LLM-based systems of agents trained with multi-agent reinforcement learning~\citep{motwani_secret_2024,mathew_hidden_2024,anwar_foundational_2024}.
In these settings, emergent communication may happen incidentally.
Disconcertingly, as~\cite{mathew_hidden_2024} show, the communication systems that develop may appear human readable, but secretly be carrying alternative meanings.
This raises the issue of decoding such covert languages, which may be possible using the frameworks and methods presented here.

\section{Conclusion}

We introduced the problem of decoding communication systems under partial observability.
Firstly, we looked at how the relationship between communication strategies and environment-level strategies complicates the problem of decoding the communication protocol.
The problem was formally analysed by defining the concept of environment-level and communicative strategic equivalence classes.
We saw to perform inference, we needed to assume that the agents in the target community are \textit{rational}, however, there may be many optimal policies that are consistent with the observed data.
We introduced a learning algorithm composed of three components: a joint policy, a transition model, and a state decoder, and evaluated them in a simple gridworld problem.

The results of this work are preliminary and there are many avenues for future work.
The most immediate is to evaluate the performance of the state decoder in higher dimensional domains.
The analysis and empirical evaluations were conducted in a simple environment, so it is unclear how well this method will generalise to more complex environments.
In our setting, the communication system is simple --- each episode is solved by a single message from the speaker.
Additionally, the speaker and listener used in this gridworld share no common observables, which means that messages cannot have contextual meanings.
More complex communication systems will require adapting the state decoder architecture.

\footnotesize
\bibliographystyle{apalike}
\bibliography{references} 

\end{document}